\documentclass[a4paper,twoside,12pt]{article}

\usepackage{multicol} 
\usepackage{graphicx}
\usepackage{amsmath}
\usepackage{mathptmx}
\usepackage{hyperref}
\usepackage{amsthm}
\usepackage{amsfonts}
\usepackage{lipsum}
\usepackage{amssymb}

\usepackage{caption}
\usepackage{subcaption}

\usepackage{float}

\usepackage[T1]{fontenc}
\usepackage[utf8]{inputenc}

\usepackage[hmarginratio=1:1,top=32mm,columnsep=20pt]{geometry} 
\usepackage{abstract} 
\usepackage{fancyhdr}
\title{Mapping The Layers of The Ocean Floor With a Convolutional Neural Network}

\author{
Fernandes, G.G.D.$^1$; Oliveira, V.S.P.P.$^2$;  Astolfo, J.P.I.$^3$\\ 
\footnotesize $^1$Department of Physics, Instituto Superior Técnico, Lisbon, LIS, Portugal \\
\footnotesize $^2$Department of Chemistry and Physics, Federal University of Espírito Santo, Alegre, ES, Brazil \\
\footnotesize $^3$Department of Physics, Federal University of São Carlos, São Carlos, SP, Brazil \\ 
\vspace{-0mm}
}

\date{~~~}
\begin{document}

\maketitle 

\thispagestyle{fancy} 

\vspace{-2.0cm}

\begin{abstract}

\vspace{-3mm}
\noindent The mapping of ocean floor layers is a current challenge for the oil industry. Existing solution methods involve mapping through seismic methods and wave inversion, which are complex and computationally expensive. The introduction of artificial neural networks, specifically UNet, to predict velocity models based on seismic shots reflected from the ocean floor shows promise for optimising this process. In this study, two neural network architectures are validated for velocity model inversion and compared in terms of stability metrics such as loss function and similarity coefficient, as well as the differences between predicted and actual models. Indeed, neural networks prove promising as a solution to this challenge, achieving Sørensen-Dice coefficient values above 70\%.
\end{abstract}

\vspace{8mm}
\begin{multicols}{2} 
\section{Introduction}

The seismic method consists of capturing mechanical waves, triggered by artificial sources, in sensors positioned on the earth's or sea surface, on the ocean floor, and even within drilled wells \cite{felipe}. Assuming that the Earth is an acoustic medium, each sensor’s main parameter is the wave propagation speed, which varies as it passes through different surface layers. Each shot is recorded, containing characteristics relevant to soil positioning, which are later converted into an image of the geological subsurface layers.

The advantage of this technique is that it mitigates exploratory risks, as it provides prior geological knowledge of the terrain. Widely used by the oil sector to maximise economic yield and minimise environmental impacts, two-dimensional seismic images are generated and segmented to identify potential oil reservoirs. However, there is a disadvantage to this technique. To generate an accurate model from the shots, a high computational cost is required, as this is essentially an ill-posed inverse problem, highly sensitive to initial conditions.

To address this, automatic segmentation methods are used as an alternative, including Deep Learning (DL), specifically Artificial Neural Networks (ANNs). As the name suggests, these are computational techniques inspired by the central nervous system, specialised in pattern recognition. Like the human brain, their learning process relies on synaptic weights, which connect the network’s neurons, together forming a layer. As information passes through the input layers, it is distributed in parallel to all neurons, which process the values, and ultimately a signal is propagated in the output layer.

Based on this premise, our general objective was to use the ANN, UNet, for the segmentation process. The specific objectives included pre-processing seismic images through the \textit{deep wave} algorithm, training the DL models with different hyperparameters, and performing statistical comparisons using metrics, box-plot graphs, and images generated by the model in comparison with the Ground Truth (GT). The outcome was UNet models achieving Sørensen-Dice coefficient rates above 70\%, demonstrating the network’s potential for oil identification in images.

\section{Theoretical Framework}

\subsection{Artificial Neural Networks}

ANNs are computational models inspired by the functioning of the human brain, designed to learn from input data and perform specific tasks. These models consist of a set of interconnected neurons that process information and generate outputs based on mathematical rules.

The structure of an ANN is composed of layers of neurons, with each layer containing a variable number of neurons. The input layer receives the network's data, while the hidden layers process this data to produce new information. Finally, the output layer provides the network’s final response \cite{5}.

One of the main advantages of ANNs is their ability to learn from examples. During network training, various input examples are presented along with the expected output for each. The network adjusts its weights to minimise the difference between the produced and expected outputs. This process is called backpropagation learning.

There are various types of ANNs, each with its unique architecture and specific application. Examples include Convolutional Neural Networks (CNNs), used in computer vision tasks; recurrent neural networks, used in tasks involving data sequences; and feedforward neural networks, applied in classification and regression problems \cite{6}.

ANNs have been widely utilised in numerous fields, including pattern recognition, speech processing, image processing, and gaming. Furthermore, with increased computational power and access to large datasets, neural networks have proven increasingly effective in complex tasks, such as object recognition in images and automatic language translation \cite{7}.

Despite their advantages, ANNs also have some limitations. Generally, they require large amounts of training data to achieve good performance and can be highly sensitive to noise and variations in data. Additionally, the interpretability of ANNs can be challenging, as the network’s weights and internal connections are difficult for humans to understand \cite{8}.

\subsubsection{Convolutional Neural Network: UNet}

A CNN is a particular type of ANN that receives image datasets as input and produces a label as output; they can essentially be summarised as a set of convolution and \textit{pooling} layers that receive input tensors in the form (height, width, channels), excluding the batch dimension \cite{chollet}.

\begin{figure*} [ht]
    \centering
    \includegraphics[width=0.75\textwidth]{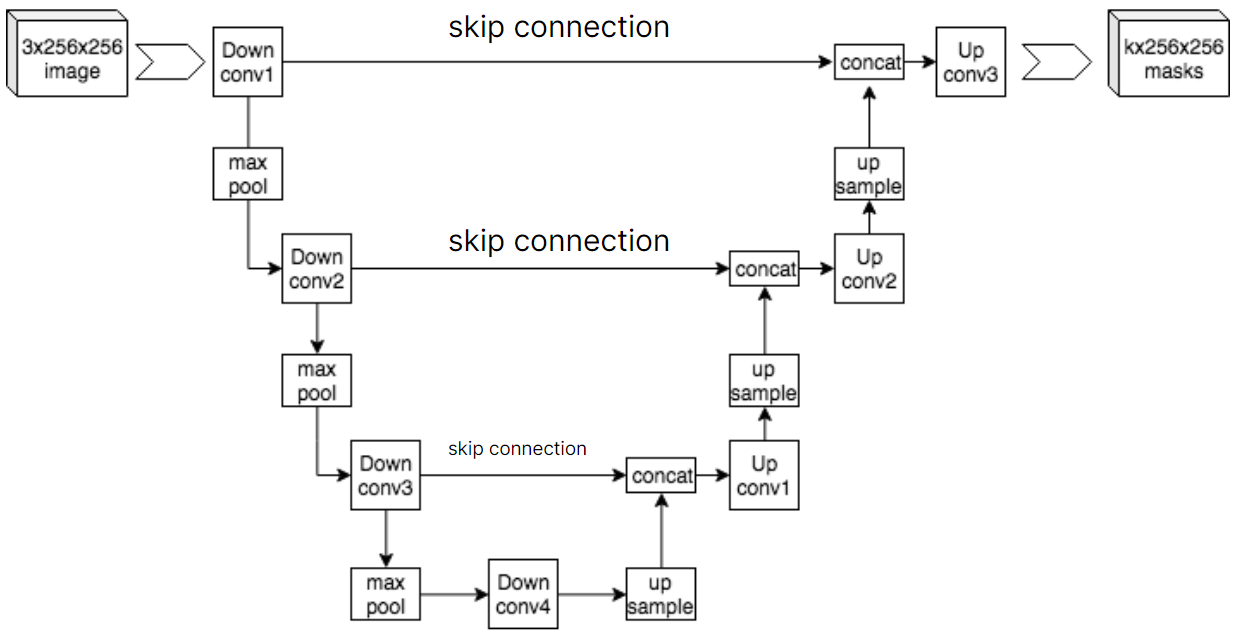}    \caption{Example of a UNet. Given the complexity of the ANNs used in this study, we present an illustration of their architectures instead.}
    \label{fig:unet}
\end{figure*}

In this study, we used a type of CNN called UNet. Developed in 2015 by Ronneberger, Fischer, and Brox \cite{ronne} and illustrated in Figure \ref{fig:unet}, it is named after the U-shape of its architecture. Initially conceived and applied to biomedical images due to its superior performance with this type of data, it is also efficient in achieving results with small datasets \cite{ronne}. The main difference between a CNN and a UNet lies in the output layer, as the former provides a label, whereas the latter outputs images.

Another significant difference is the use of \textit{skip connections}. As observed in Figure \ref{fig:unet}, these connections link the layers of the encoding and decoding blocks, allowing for the recovery of spatial information lost through \textit{pooling} layer operations \cite{drod}.

\subsection{Seismic Velocity Model Inversion Using Deep Learning Techniques}

Seismic velocity model inversion is a technique used in oil and gas exploration, aimed at inferring the internal structure of the subsurface from seismic data. The goal is to obtain a velocity model that can be used to allocate drilling wells and identify hydrocarbon reservoirs.

Traditional inversion techniques are based on optimisation methods that require defining an initial model and minimising an objective function. However, these methods are limited by the presence of noise in the seismic data and the lack of information about the geological structure of the subsurface \cite{9}.

Recently, DL-based techniques have been proposed to perform seismic velocity model inversion. These techniques use deep ANNs to learn the relationship between seismic data and the velocity model. Instead of explicitly defining an initial model and an objective function, the ANN is trained to map seismic data to a velocity model that produces outputs similar to real data.

One of the main advantages of these techniques is their ability to handle complex, noisy data. ANNs can learn complex patterns in seismic data and filter out noise, producing more accurate velocity models. Additionally, ANNs can incorporate additional geological information, such as well data and geophysical information, to produce more accurate models \cite{10}.

One of the most commonly used DL techniques for seismic velocity model inversion is the convolutional neural network. This technique uses convolutional layers to extract features from seismic data and regression layers to produce the velocity model. During training, the neural network is adjusted to minimise the difference between the produced velocity model and the real model.

Another DL technique used in seismic velocity model inversion is the generative adversarial network. This technique uses two neural networks, a generator and a discriminator, which are trained together. The generator network produces synthetic velocity models, while the discriminator network evaluates the quality of these models. During training, the generator network is adjusted to produce models that deceive the discriminator network and are similar to real models.

DL-based seismic velocity model inversion techniques have shown promising results in experimental studies and practical applications. However, there are still challenges to overcome, such as the interpretability of the models produced by ANNs and the need for large datasets to train the networks \cite{11}.

In summary, ANNs are a powerful tool for solving complex machine learning problems. With their ability to learn from examples and adapt to different data types, they have become increasingly relevant in various fields. However, much remains to be explored in terms of the interpretability and limitations of these models.

\section{Materials and Methods}

For training the ANN, the input data consisted of the shots generated by sensors, and the output data were the models. The seismic data were generated using \textit{DeepWave}, a wave propagation simulation package, which allows for forward modelling and backpropagation to calculate the gradient of the outputs relative to the inputs (thus enabling inversion/optimisation)~\cite{deepwave}.

Next, the hyperparameters for the two UNet models used were selected. The main difference between the two networks lies in their final layer, where one does not include a \textit{skip connection} between its first and last layer. Accordingly, they are named UNet and UnetMod, respectively.

All documentation was developed in the Python programming language on the Google Colaboratory platform, using the TensorFlow.Keras interface for building the AI models.

\subsection{Datasets}

The dataset for training and validating the network consists of seismic shots simulated from velocity models, which were also simulated. The \textit{DeepWave} package was used to generate the seismic shots. Each set of shots serves as an input to the ANN. To assess the influence of dataset size on the network's efficiency, two datasets were created: the first with 300 simulated elements and the second with 1,000 simulations. All generated velocity models and seismic shots form a data grid of 128 x 128 dimensions.

\subsubsection{Model Generator}

\begin{figure*} [ht]
    \centering
    \includegraphics[width=0.75\textwidth]{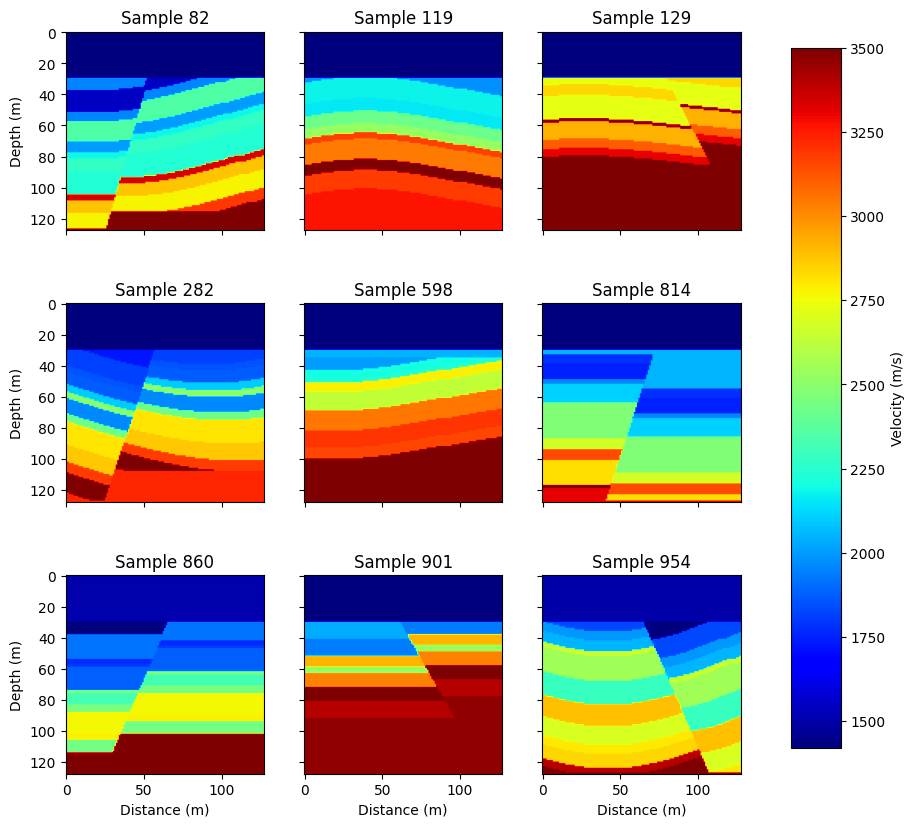}    \caption{Samples of generated velocity models.}
    \label{fig:models}
\end{figure*}

The velocity models are generated considering different geological features, such as faults, erosion processes, fractures, and the range of minimum and maximum velocities across different propagation media. Two sets of velocity models are created. The first set is based on simple models without geological defects, with fewer simulated elements, and is considered in the scope of this study as the simple set. The second set includes geological defects, contains more simulated elements, and presents greater complexity, thus regarded as the complex set. Figure~\ref{fig:models} presents samples of the generated models, showing more complex geological features.

\subsubsection{Shot Generator}

The seismic shots used as input for the networks employ the \textit{DeepWave} package, which implements wave propagators as PyTorch modules. Its functions enable wave modelling and backpropagation in the ANN to calculate the gradient of the network outputs relative to the inputs. Various shot parameters can be configured in this package, such as the arrangement of simulated wave receivers (number, position, spacing, among other parameters) and wave frequency.

\subsection{Google Colaboratory, TensorFlow, and \\Supercomputer}

This project was initially developed on \textit{Google Colaboratory}. This platform allows Python code to be executed directly from Google’s cloud servers, where one of its main advantages is the use of its Graphics Processing Units (GPU). In combination, for developing the ANNs, we used the Keras interface, a high-level API independently built on TensorFlow. TensorFlow is a library capable of splitting training into multiple parts and executing them in parallel across several GPUs, enabling the training of colossal neural networks on enormous training datasets by distributing calculations across hundreds of servers within a reasonable timeframe \cite{geron}.

\subsection{Artificial Neural Network Training}

For training the ANN, the cross-validation technique was applied to assess the model’s generalisation capacity in different contexts. The evaluation metric used was the Mean Square Error (MSE), which calculates the average squared difference between observed and predicted values. The dataset was randomly divided into 10 batches, maintaining a split of 64\% for training and 16\% for validation. Based on this set, 10 random batch iterations were performed. The result is the simple average of the metric outcome across all iterations.

For the testing phase, the remaining 20\% was used. This phase was responsible for predicting new results and, consequently, testing the network's performance on new data. The Sørensen–Dice Coefficient (DSC) was chosen as the metric, designed to assess the similarity rate between two images. It was selected for its intuitive visualisation, as it provides values between 0.0 and 1.0, which can be easily converted into percentages.

To optimise the training, a preliminary selection of hyperparameters was made. Initially, no direct focus was placed on these parameters, as callback functions were used to monitor the training. The primary focus was on batch size, consisting of small sample sets processed simultaneously by the model to facilitate GPU memory allocation \cite{chollet}. Empirically, a batch size of 8 was chosen for both architectures.

\subsubsection{Statistical Analysis}

To visually represent the DSC metric, box-plot graphs were used to assess the stability of the models. Additionally, a visual comparison was performed between the Ground Truth (GT) and the predicted images to identify potential prediction biases, such as the presence of artifacts and other spurious signals.

\section{Results and Discussion}

\begin{figure*}[htbp]
    \centering
    \includegraphics[width=0.7\textwidth]{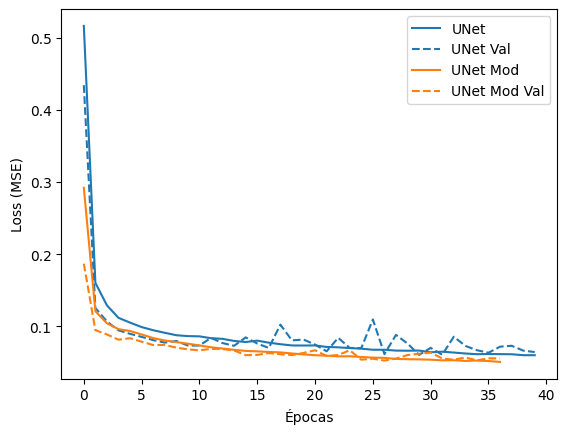}
    \caption{Loss curves for the two trained network architectures.}
    \label{fig:loss}
\end{figure*}

\begin{figure*}
    \centering
    \begin{subfigure}[b]{\textwidth}
         \centering
         \includegraphics[width=0.8\textwidth]{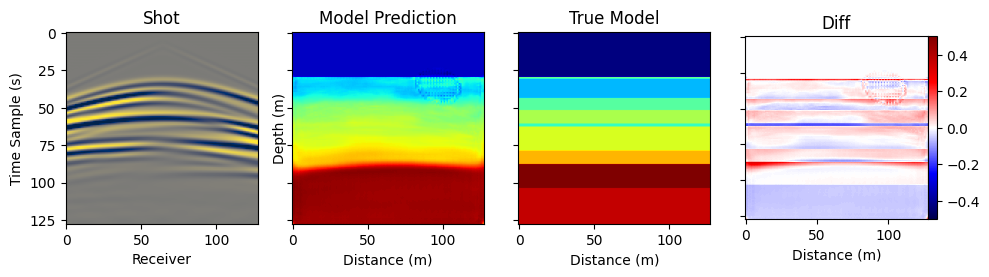}
         \caption{Simple sample, without geological faults.}
         \label{fig:simple}
     \end{subfigure}
     \begin{subfigure}[b]{\textwidth}
         \centering
         \includegraphics[width=0.8\textwidth]{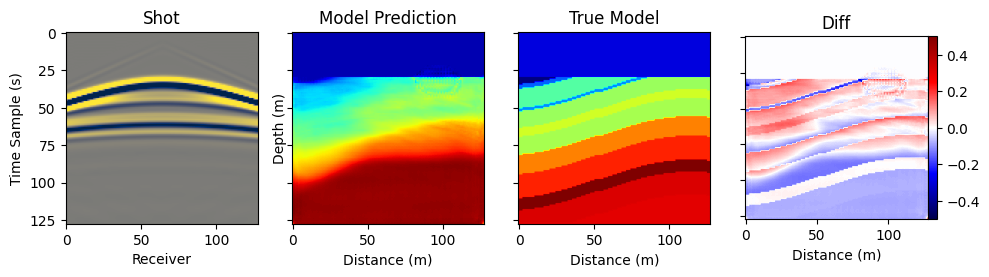}
         \caption{Sample with folding.}
         \label{fig:dobra}
     \end{subfigure}
     \begin{subfigure}[b]{\textwidth}
         \centering
         \includegraphics[width=0.8\textwidth]{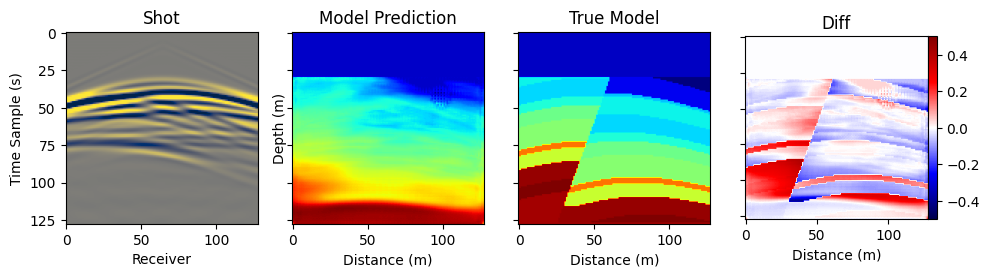}
         \caption{Sample with geological fault and folding.}
         \label{fig:falha}
     \end{subfigure}
     
    \caption{Comparison between the network's seismic shot input, the predicted models, actual models, and the difference (subtraction) between the predicted and actual models.}
    \label{fig:compare}
\end{figure*}

\begin{figure*}
    \centering
    \includegraphics[width=0.7\linewidth]{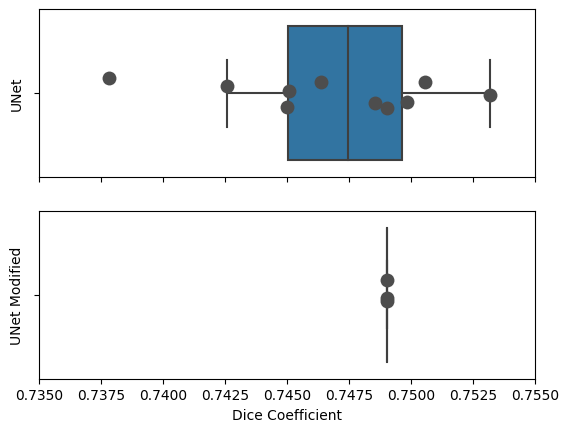}
    \caption{Sørensen-Dice coefficient for the trained networks, an indicator of the stability of similarity between actual and predicted models by the networks.}
    \label{fig:dice}
\end{figure*}

Figure \ref{fig:loss} presents the loss function curves for each architecture trained in this study. It is notable that the modified UNet displays a more stable validation curve around the training curve when compared to the unmodified UNet. The behaviour of the validation curves indicates that the solution does not exhibit overfitting or underfitting issues. Both networks require approximately the same time for their loss functions to cease significant variations.

It is interesting to note that the models predicted by the modified UNet, when compared with the actual models, exhibit a circular artifact consistently positioned in the upper right region of the images. Figure~\ref{fig:compare} shows the model comparison, with the circular artifact easily visualised in the difference image. The dataset used tends to contain a geological fault near the region where the artifact appears, and the network generalises this element in all predicted models.

In general, the networks demonstrate satisfactory predictions when the velocity model is simpler (Figure~\ref{fig:simple}) and with fewer geological faults. Velocity models containing folds (Figure~\ref{fig:dobra}) are also satisfactorily predicted by the ANN, although the information about the layer interface is lost due to the predicted gradients. Models with folds combined with a geological fault (Figure~\ref{fig:falha}) exhibit lower accuracy in network prediction. In these more complex cases, the predicted models are capable of identifying the folds, but the faults become ambiguous. Nonetheless, it is possible to observe a basic identification of these geological features, and further training of the network with more epochs and a larger dataset containing more samples with faults and folds could enhance the identification of these structures.

It is possible to analyse the similarity between the images of the models predicted by the network and the actual models used as input for generating the seismic shots. Figure~\ref{fig:dice} presents the Sørensen-Dice coefficient for each fold during network cross-validation, for the two architectures used. It can be noted that the similarity obtained by the modified UNet exhibits greater stability in the similarity coefficient across the folds, whereas the unmodified UNet is less stable.

The similarity of approximately 75\% highlights the differences between the images caused by the presence of gradients in the predicted models, as the layer boundaries of the models used as input for simulating seismic shots are well-defined and their interfaces are discontinuous.

Network optimisations can be made with small changes to the architectures or the generation of seismic shots from the velocity model, so that the prediction becomes more accurate. The dataset itself is a limiting factor for network training and validation, and increasing the number of input models should improve network efficiency. Nevertheless, the solutions produced by the network prove to be stable.

\section{Conclusion}

ANNs prove to be promising solutions for the challenge of velocity model inversion and ocean floor mapping; satisfactory results were achieved regardless of the UNet architecture used. In future work, we intend to investigate the circular artifacts that appear in some predictions. Another next step would be to introduce a \textit{Physics Informed} stage to optimise the network for complex models; concurrently, generating more training data to enhance the generalisation capability of ANNs.

\section{Acknowledgements} 

We extend our thanks to the Brazilian Centre for Physics Research for the invaluable opportunity to participate in the sixth edition of the CBPF Advanced School of Experimental Physics, an academically, interpersonally, and professionally enriching experience. In particular, we thank the professors and monitors of the Applied Artificial Intelligence module.

\end{multicols}
\end{document}